\documentclass[sigconf]{acmart}
\usepackage{color}
\usepackage{fancyhdr}
\usepackage[ruled,vlined]{algorithm2e}

\AtBeginDocument{%
  \providecommand\BibTeX{{%
    \normalfont B\kern-0.5em{\scshape i\kern-0.25em b}\kern-0.8em\TeX}}}

\usepackage{url}
\usepackage{soul, hyperref, xcolor, enumitem}
\usepackage[noend]{algpseudocode}
\usepackage{balance}
\newcommand{\ModelName}{\textsc{PSN}}
\usepackage{amsmath,amsthm,mathtools}
\usepackage{latexsym,graphicx,multirow,booktabs}
\usepackage{makecell}

\usepackage{subfigure}
\usepackage{microtype}

\newcommand{\KL}{\mathbb{D}_{\rm{KL}}}

\copyrightyear{2021}
\acmYear{2021}
\setcopyright{acmcopyright}\acmConference[SIGIR '21]{Proceedings of the 44th International ACM SIGIR Conference on Research and Development in Information Retrieval}{July 11--15, 2021}{Virtual Event, Canada}
\acmBooktitle{Proceedings of the 44th International ACM SIGIR Conference on Research and Development in Information Retrieval (SIGIR '21), July 11--15, 2021, Virtual Event, Canada}
\acmPrice{15.00}
\acmDOI{10.1145/3404835.3462995}
\acmISBN{978-1-4503-8037-9/21/07}

\begin{document}
\fancyhead{}
\title{Pseudo Siamese Network for Few-shot Intent Generation}

\author{Congying Xia}
\email{cxia8@uic.edu}
\affiliation{%
  \institution{University of Illinois at Chicago}
  \city{Chicago}
  \state{IL}
  \country{USA}
}

\author{Caiming Xiong}
\email{cxiong@salesforce.com}
\affiliation{%
  \institution{Salesforce Research}
  \city{Palo Alto}
  \state{CA}
  \country{USA}
}

\author{Philip Yu}
\email{psyu@uic.edu}
\affiliation{%
  \institution{University of Illinois at Chicago}
  \city{Chicago}
  \state{IL}
  \country{USA}
}


\begin{abstract}
Few-shot intent detection is a challenging task due to the scare annotation problem. In this paper, we propose a Pseudo Siamese Network (PSN) to generate labeled data for few-shot intents and alleviate this problem. PSN consists of two identical subnetworks with the same structure but different weights: an action network and an object network. Each subnetwork is a transformer-based variational autoencoder that tries to model the latent distribution of different components in the sentence. The action network is learned to understand action tokens and the object network focuses on object-related expressions. It provides an interpretable framework for generating an utterance with an action and an object existing in a given intent. Experiments on two real-world datasets show that PSN achieves state-of-the-art performance for the generalized few shot intent detection task.
\end{abstract}

\ccsdesc[500]{Computing methodologies~Artificial intelligence}
\ccsdesc[500]{Computing methodologies~Natural language processing}
\ccsdesc[500]{Computing methodologies~Information extraction}

\keywords{Few-shot Learning, Intent Detection, Text Generation}
\maketitle

\section{Introduction}
Intelligent assistants have gained great popularity recently. Companies are striving to deliver their products either on speaker devices such as Amazon Alexa, or smartphones such as Siri from Apple. To provide an intelligent conversational interface, these assistants need to understand the user's input correctly. Among all the natural language understanding tasks, intent detection is an important and essential one. It aims at understanding the goals underlying input utterances and classifying these utterances into different types of intents. For example, given an input utterance, ``How's the weather in Chicago tomorrow?'', the system needs to identify the intent is query weather.
With the development of deep learning techniques, intent detection has achieved great success by formalizing it as a text classification task under the supervised learning paradigm \cite{xu2013convolutional,chen2016end}. These works rely on a large amount of labeled data to train the intent detection model.
Certain restrictions like requiring sufficient labeled examples for each class limit these models' ability to adapt to previously unseen classes promptly. Recently, researchers are interested in achieving decent performance with reduced human annotation and extending models' ability to detect new classes. Low-resource learning paradigms \cite{9319339, zhang2020discriminative} like Zero-shot learning \cite{xia2018zero} and Few-shot learning \cite{xia2020composed, nguyen2020semantic, xia2021incremental} have drawn a lot of attention recently. In this work, we focus on the task of identifying few-shot intents which only have a few labeled examples.

The bottleneck for identifying few-shot intents is the lack of annotations. If we can generate high-quality pseudo-labeled examples for these few-shot intents, we can effectively alleviate this issue and improve the performance. There are only a few previous works \cite{wei2019eda,malandrakis2019controlled,yoo2019data, liu2021augmenting} that try to augment the training data with generation methods and alleviate the scarce annotation problem. 
However, these models utilize simple neural networks with limited model capacity, like LSTMs \cite{hochreiter1997long}, to do text generation. Furthermore, these methods do not consider the inner structure for an intent. Naturally, an intent can be defined as an action with an object \cite{xu2019open}. For example, the intent of the input ``wake me up at 7 am'' is to set an alarm. This intent consists of an action “Set" and an object ``Alarm''. In this paper, we propose a Pseudo Siamese Network (PSN) that generates labeled examples for few-shot intents considering the inner structure of an intent.
PSN consists of two identical subnetworks with the same structure but different weights: an action network and an object network.
To utilize the powerful pre-trained language models and capture the latent distribution of sentences with different intents, we propose to use transformer-based \cite{vaswani2017attention} variational autoencoders \cite{kingma2013auto} as the sub-networks to model different components in the sentences. 
The action network is learned to understand action tokens and the object network focuses on object-related expressions. 
During the inference, PSN generates an utterance with a given intent by controlling the action generation and the object generation separately in two subnetworks. It provides an interpretable framework for generating an utterance with an action and an object existing in a given intent. 

To quantitatively evaluate the effectiveness of PSN for augmenting training data in low-resource intent detection, experiments are conducted for the generalized few-shot intent detection task (GFSID) \cite{xia2020cg}. GFSID is a more practical setting for few-shot intents. It not only considers the few-shot intents with a few labeled examples, but also includes existing intents with enough annotations.
Formally, GFSID aims to discriminate a joint label space consisting of both existing many-shot intents and few-shot intents. In summary, the main contributions of our work are as follows. 1) We propose a Pseudo Siamese Network to generate high-quality labeled data for few-shot intents and alleviate the scarce annotation problem. 2) PSN provides an interpretable framework for generating an utterance with an action and an object belonging to a given intent by controlling each part in a subnetwork. 3) Empirical experiments conducted on two real-world datasets show the effectiveness of our proposed model on the generalized few-shot intent detection task.

\begin{figure*}[h]
    \centering
    \includegraphics[width=\linewidth]{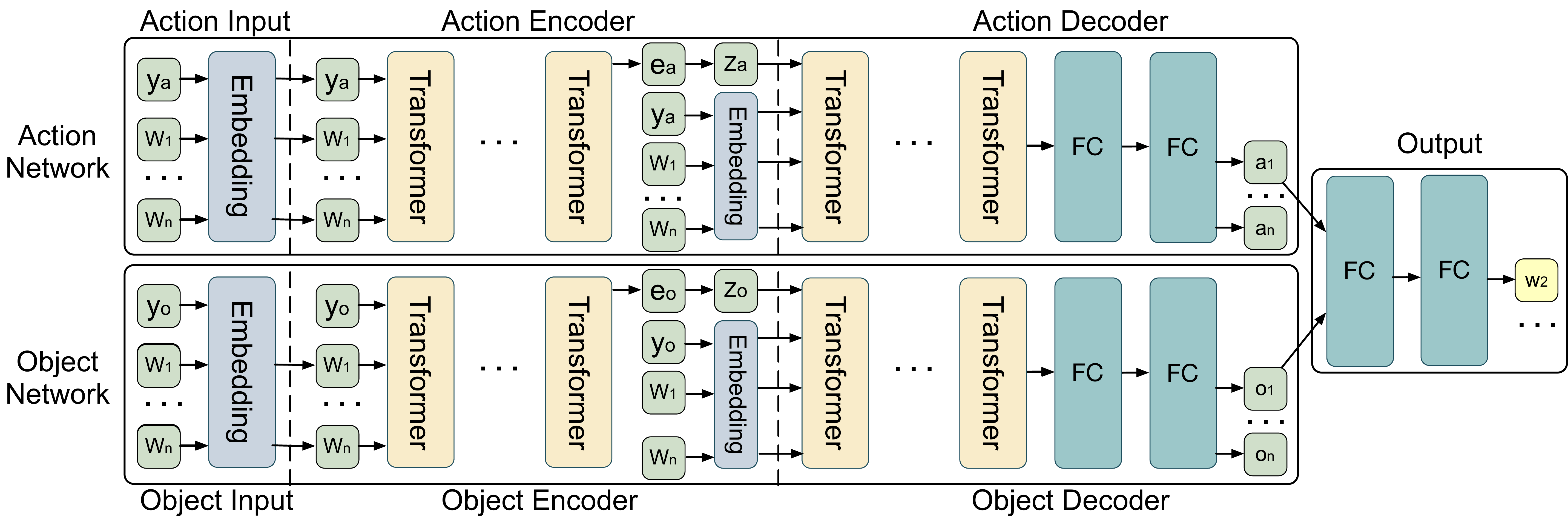}
    \vspace{-0.3in}
    \caption{The overall framework of Pseudo Siamese Network. FC is short for Fully-Connected layers.}
    \vspace{-0.1in}
    \label{fig:framework}
\end{figure*}

\section{Pseudo Siamese Network}
In this section, we introduce the details for the proposed Pseudo Siamese Network (PSN). 
As illustrated in Figure \ref{fig:framework}, PSN consists of two identical subnetworks: an action network and an object network.
These two subnetworks have the same structure with different weights. Each subnetwork is utilized to model different components in the utterances. The action network is used to learn action tokens and the object network is proposed to focus on object-related expressions.
Specifically, each subnetwork is a transformer-based variational autoencoder that consists of an encoder and a decoder. Each encoder and decoder are a stack of multiple transformer layers.

\subsection{Input Representation}
\label{input}
Each training instance consists of an input sentence and a corresponding intent. To capture the inner structure of the intent, we define the intent as a pair of an action $y_a$ and an object $y_o$. Given an input sentence $s = (w_1, w_2, ..., w_n)$ with $n$ tokens, we construct two text pairs and feed them separately into two subnetworks. We feed the action token together with the input sentence into the action network, while the object token and the input sentence are fed into the object network.
To formalize the input for transformer-based models, we add a special start-of-sequence ([CLS]) token at the beginning of each input and a special end-of-sequence ([SEP]) token at the end of each sequence.

Formally, the input for the action network is formatted as ([CLS], $y_a$, [SEP], $w_1$, $w_2$, ..., $w_n$, [SEP]) and the input for the object network is ([CLS], $y_o$, [SEP], $w_1$, $w_2$, ..., $w_n$, [SEP]). The input of each subnetwork consists of two sentences. In this paper, we refer ([CLS], $y_a$, [SEP]) and ([CLS], $y_o$, [SEP]) to as $S_1$, and ($w_1$, $w_2$, ..., $w_n$, [SEP]) as $S_2$ in each subnetwork. For each input in the subnetwork, they are tokenized into subword units by WordPiece \cite{wu2016google}. The input embeddings of a token sequence are represented as the sum of three embeddings: token embeddings, position embeddings \cite{vaswani2017attention}, and segment embeddings \cite{devlin2019bert}. These embeddings for input representation are shared between the action network and the object network.

\subsection{Network Structure}
The overall framework of Pseudo Siamese Network is illustrated in Figure \ref{fig:framework}. PSN consists of an action network and an object network. The action network has an action encoder and an action decoder while the object network has an object encoder and an object decoder. We will describe the encoders and the decoders separately in this section.

\subsubsection{Encoders}
Two encoders including the action encoder and the object encoder are contained in PSN. The action encoder encodes the action and the input sentence into a latent variable $z_a$ while the object encoder encodes the object and the input sentence into a latent variable $z_o$. Multiple transformer layers \cite{vaswani2017attention} are utilized in the encoders. Each transformer layer models the self-attentions among all the tokens.
For the $l$-th transformer layer, the output of a self-attention head $A_l$ is computed via:
\begin{equation}
{A_l} = \text{softmax} \left(\frac{QK^{\top}}{\sqrt {d_k}}  \right){V}, 
\end{equation}
where $Q, K, V$ are queries, keys, and values projected from the previous layer $H^{l-1}$ and parameterized by matrices $W^l_Q, W^l_K, W^l_V{\in}\mathbb{R}^{d_h{\times}d_k}$:
\begin{align}
    Q = H^{l-1}W^l_Q, \;
    K = H^{l-1}W^l_K, \;
    V = H^{l-1}W^l_V. \;
\end{align}

The embeddings for the [CLS] token that output from the last transformer layer in the encoder are used as the encoded sentence-level information. The encoded sentence-level information is denoted as ${e}_a$ in the action encoder and ${e}_o$ in the object encoder. ${e}_o$ and  ${e}_a$ are encoded into $z_a$ and $z_o$ to model the distribution for the action and the object separately.

By modeling the true distributions, $p(z_a|x, y_a)$ and $p(z_o|x, y_o)$, using a known distribution that is easy to sample from \cite{NIPS2014_5352}, we constrain the prior distributions, $p(z_a|y_a)$ and $p(z_o|y_o)$, as multivariate standard Gaussian distributions. A reparametrization trick \cite{kingma2013auto} is used to generate the latent vectors $z_a$ and $z_o$ separately. Gaussian parameters ($\mu_a$, $\mu_o$, $\sigma_a^2$, $\sigma_o^2$) are projected from ${e}_a$ and ${e}_o$:
\begin{align}
    \mu_a  = {e}_a{{W}_{\mu_a} } + {b_{\mu_a} }, \\
    {\log(\sigma_a ^2)} = {e}_a{{W}_{\sigma_a} } + {b_{\sigma_a} }, \\
    \mu_o  = {e}_o{{W}_{\mu_o} } + {b_{\mu_o} }, \\
    {\log(\sigma_o ^2)} = {e}_o{{W}_{\sigma_o} } + {b_{\sigma_o} }, 
\end{align}
where ${W}_{\mu_a}, {W}_{\mu_o}, {W}_{\sigma_a}, {W}_{\sigma_o} \in \mathbb{R}^{d_h \times d_h}$ and $b_{\mu_a}, b_{\mu_o}, b_{\sigma_a}, b_{\sigma_o} \in \mathbb{R}^{d_h}$. Noisy variables $\varepsilon_a \sim \mathcal{N}(0, \mathrm{I}), \varepsilon_o \sim \mathcal{N}(0, \mathrm{I})$ are utilized to sample $z_a$ and $z_o$ from the learned distribution:
\begin{equation}
	{z_a} = \mu_a + {\sigma_a} \cdot \varepsilon_a,
	{z_o} = \mu_o + {\sigma_o}\cdot \varepsilon_o. 
\end{equation}

\subsubsection{Decoders}
The decoder utilizes latent variables together with labels to reconstruct the input sentence $p(s|z_a, z_o, y_a, y_o)$. As shown in Figure \ref{fig:framework}, the action decoder takes $z_a$, $y_a$, and the sentence $s = (w_1, w_2, ..., w_n)$ as the input while the input of the object decoder are $z_o$, $y_o$, and the sentence $s$. The label components ($y_a$, $y_o$) and the sentence $s$ are embedded with an embedding layer. The embedding parameters are shared with the input representation.

To keep a fixed length and introduce the latent information $z_a$ and $z_o$ into the decoders, we replace the first [CLS] token with $z_a$ and $z_o$ in each sub-network. The decoders are also built with multiple transformer layers. Text generation is a sequential process that uses the left context to predict the next token. Inspired by \cite{dong2019unified} that utilizes specific
self-attention masks to control what context the prediction conditions on, we apply the sequence-to-sequence attention mask proposed in Dong\shortcite{dong2019unified} in the decoders to simulate the left-to-right generation process.
With the attention mask applied in the decoders, tokens in $S_1$ can only attend to tokens in $S_1$, while tokens in $S_2$ can attend to tokens in $S_1$ and all the left tokens in $S_2$. For the first tokens in two decoders, $z_a$ and $z_a$, which hold latent information, they are only allowed to attend to themselves due to the vanishing latent variable problem. The latent information can be overwhelmed by the information of other tokens when adapting VAE to natural language generators \cite{zhao2017learning}.

After the transformer layers in the decoders, we can obtain the embedding outputs for these two sequences: ($z_a$, $y_a$, [SEP], $w_1$, ..., $w_n$, [SEP]) and ($z_o$, $y_o$, [SEP], $w_1$, ..., $w_n$, [SEP]). To further increase the impact of the latent information and alleviate the vanishing latent variable problem, we concatenate the output embeddings of $z_a$ to other token embeddings in the first sequence and concatenate $z_o$ to other token embeddings in the second sequence. The hidden dimension increases to $2 \times d_h$ after the concatenation. To reduce the hidden dimension to $d_h$ and get the embeddings to decode the vocabulary, two fully-connected (FC) layers followed by a layer normalization \cite{ba2016layer} are applied on top of the transformer layers. Gelu is used as the activation function in these two FC layers. For the token at position $i$ in the sentence $s$, the output representation from the action decoder is denoted as $a_i$ and $o_i$ from the object decoder.

As shown in the output box of Figure \ref{fig:framework}, the outputs from action decoder and object decoder are fused together to predict the next token. An FC layer is used to fuse these outputs:
\begin{equation}
      {m}_{i+1} = g({a}_{i}{W}_a + {o}_{i}{W}_o  + b),
\end{equation}
where ${W}_o, {W}_a \in \mathbb{R}^{d_h \times d_h}$ and $b \in \mathbb{R}^{d_h}$ are parameters, and $g$ is the GELU activation function. The fused embeddings ${m}_{i+1}$ are used to predict the token at position $i+1$ with another FC layer. The inference process iteratively decodes the output till the [SEP] token is generated.

\subsection{Loss Function}
 In the model, the loss function consists of two parts: the KL-divergence that regularize the prior distributions for two latent variables to be close to the Gaussian distributions and the reconstruction loss:
\begin{align}
	\begin{aligned}
        \mathcal{L}=  -&{\mathbb{E}_{q(z_a|x,y_a),q(z_o|x, y_o)}}[ {\log p( {x|z_a, z_o, y_a, y_o})}] \\
        +& \KL [ {q(z_a|s, y_a), p(z_a|y_a)}] \\
        +& \KL [ {q( {z_o|s,y_o}), p(z_o|y_o)}].
\end{aligned}
\end{align}

In the inference, utterances for few-shot intents are generated by sampling two latent variables, $z_a$ and $z_o$, separately from multivariate standard Gaussian distributions. Beam search is applied to do the generation. To improve the diversity of the generated utterances, we sample the latent variables for $s$ times and save the top $k$ results for each time.
These generated utterances are added to the original training datasets to alleviate the scare annotation problem.
\begin{table}[ht!]
\centering
\resizebox{\linewidth}{!}{
\begin{tabular}{l|c|c}
\toprule
 Dataset & SNIPS-NLU & NLUED \\ \midrule
Vocab Size & 10,896 & 6,761\\ 
\#Total Classes & 7 & 64 \\ 
\#Few-shot Classes & 2 & 16\\
\#Few-shots / Class & 1 or 5 & 1 or 5 \\
\#Training Examples & 7,858 & 7,430\\ 
\#Training Examples / Class & 1571.6 & 155\\
\#Test Examples & 2,799 & 1,076\\ 
Average Sentence Length   &  9.05    & 7.68\\
\bottomrule
\end{tabular}
}
\caption{Data Statistics for SNIPS-NLU and NLUED. \#Few-shot examples are excluded in the \#Training Exampels. For NLUED, the statistics is reported for KFold\_1.}
\vspace{-0.2in}
\label{dataset}
\end{table}

\begin{table*}[ht]
\centering
\resizebox{\linewidth}{!}{
\begin{tabular}{l|ccc|ccc}
\toprule
 & Seen  & Unseen & H-Mean  & Seen  & Unseen & H-Mean \\  
                     & \multicolumn{3}{c|}{SNIPS-NLU 5-shot}     & \multicolumn{3}{c}{NLUED 5-shot}    \\ \midrule 
BERT-PN  & 	95.96 $\pm$ 1.13&	86.03 $\pm$	2.00&	90.71 $\pm$ 1.19 & 83.41 $\pm$ 2.62&	60.28 $\pm$ 4.19&	69.93 $\pm$ 3.49  \\ 
BERT   &  98.34 $\pm$ 0.10& 81.82 $\pm$ 6.16 &89.22 $\pm$ 3.74&\textbf{94.12 $\pm$ 0.89}& 51.69 $\pm$ 3.19 &	66.67 $\pm$ 2.51  \\
BERT + SVAE & \textbf{98.34} $\pm$ \textbf{0.06} & 82.10 $\pm$ 4.06&	89.49 $\pm$ 2.47  & 93.60 $\pm$ 	0.63&	54.03 $\pm$ 3.91 &  68.42 $\pm$ 3.06\\
BERT + CGT  &98.32 $\pm$ 0.14&82.65 $\pm$ 4.31&89.78 $\pm$ 2.83& 93.61 $\pm$ 0.63&54.70 $\pm$ 4.06&68.96 $\pm$ 3.17  \\
BERT + EDA & 98.09 $\pm$ 0.18 &82.00 $\pm$ 3.47&89.30 $\pm$ 2.12 &93.71 $\pm$ 0.64&57.22 $\pm$ 4.35 &70.95 $\pm$ 3.35\\
BERT + {CG-BERT} &  98.30 $\pm$ 0.17& 86.89 $\pm$ 4.05 & 92.20 $\pm$ 2.32 & 93.80 $\pm$ 0.60 & 61.06 $\pm$ 4.29 & 73.88 $\pm$ 3.10\\ \midrule
BERT + {\ModelName}  &	98.16 $\pm$ 0.12&	\textbf{88.17 $\pm$ 1.19}& 	\textbf{92.89  $\pm$ 0.67}& 92.82 $\pm$ 0.90&		\textbf{64.16 $\pm$ 3.94} &		\textbf{75.81 $\pm$ 2.87}\\
\bottomrule
\end{tabular}
}
\caption{Generalized few shot experiments with 5-shot setting on SNIPS-NLU and NLUED.}
\vspace{-0.15in}
\label{exp}
\end{table*}

\begin{table*}[h]
\centering
\resizebox{0.8\linewidth}{!}{
\begin{tabular}{l|l}
\toprule
\multicolumn{2}{c}{\textbf{Query Alarm}} \\ \midrule
R1: what time is my alarm set for & G1: is my alarm set for \textcolor{red}{seven am} 
\\
R2: what time is my alarm set for tomorrow morning & G2: \textcolor{red}{tell me the alarm for saturday} morning \\
R3: tell me when it is five pm (\textbf{Set Alarm}) & B3: \textcolor{blue}{tell me when it is five pm} \\ \toprule
\multicolumn{2}{c}{\textbf{Recommendation Events}} \\ \midrule 
R4: show latest events around new york & G4: \textcolor{red}{what ' s the show} around new york \\
R5: what are all the event in area & G5: \textcolor{red}{check for} all the event \\
R6: is there anything to do tonight & B6: \textcolor{blue}{what show} is there anything to do tonight \\\bottomrule
\end{tabular}
}
\caption{Generation examples from {\ModelName}. Rs are real examples, Gs are good generation examples and Bs are bad cases.}
\vspace{-0.2in}
\label{case}
\end{table*}

\vspace{-0.15in}
\section{Experiments}
\subsection{Datasets}
To demonstrate the effectiveness of our proposed model, we evaluate {\ModelName} on two real-word datasets for the generalized few-shot intent detection task: SNIPS-NLU \cite{coucke2018snips} and NLU-Evaluation-Data (NLUED) \cite{XLiu.etal:IWSDS2019}. These two datasets were collected to benchmark the performance of natural language understanding services offering customized solutions. Dataset details are illustrated in Table \ref{dataset}.

\vspace{-0.05in}
\subsection{Baselines}
We compare the proposed model with five baselines. 1) Prototypical Network \cite{snell2017prototypical} (PN) is a distance-based few-shot learning model. BERT-PN is a variation of PN by using BERT as the encoder, which is referred to as BERT-PN.
2) BERT. We over-sampled the few-shot intents for this baseline.
3) SVAE \cite{bowman2015generating} is a variational autoencoder built with LSTMs.
4) CGT \cite{hu2017toward} adds a discriminator based on SVAE to classify the sentence attributes.
5) EDA \cite{wei2019eda} uses simple data augmentations rules for language transformation. We apply three rules in the experiment, including insert, delete and swap.
6) CG-BERT \cite{xia2020cg} is the first work that combines CVAE with BERT to do few-shot text generation. BERT is fine-tuned with the augmented training data for these generation baselines. The whole pipelines are referred to as BERT + SVAE, BERT + CGT, BERT + EDA and BERT + CG-BERT in Table \ref{exp}.

For {\ModelName}, we use the first six layers in BERT-base to initialize the weights in the encoders transformer layers while the latter six layers are used to initialize the decoders. {\ModelName} is trained with a learning rate equal to 1e-5 in 100 epochs and each epoch has 1000 steps. The batch size is 16. New utterances are generated by sampling the latent variables $s=10$ times and choosing the top $k=30$ utterances.

\vspace{-0.18in}
\subsection{Results}
For SNIPS-NLU, the performance is calculated with the average and the standard deviation over 5 runs. The results on NLUED are reported over 10 folds.
Three metrics are used to evaluate the model performances, including the accuracy on existing intents (Seen), the accuracy on few-shot intents (Unseen) together with their harmonic mean (H-mean) \cite{xia2020cg}. The harmonic mean is high only when the accuracy on both existing intents (Seen) and few-shot intents (Unseen) are high.
As illustrated in Table \ref{exp}, {\ModelName} achieves state-of-the-art performance on Unseen accuracy and H-mean and comparable performance on Seen accuracy. Compared to the few-shot learning baseline, BERT-PN, {\ModelName} improves the F1 score by 2.4\% from 90.71\% to 92.89\% for the NULED 5-shot setting. Compared to other data augmentation baselines, we improve the best baseline CG-BERT by 2.6\% from 73.88\% to 75.81\%. The improvement mainly stems from the high quality of the generated examples for few-shot intents, which leads to significantly increased Unseen accuracy and H-mean.

To evaluate the quality of the generated utterances and interpret how can {\ModelName} generate examples for few-shot intents, we show some examples generated by {\ModelName}. As illustrated in Table \ref{case}, {\ModelName} generates good examples by providing new slot values either for objects or new words for actions. For example, G1 generates ``seven am'' for the alarm object and G4 provides ``the show'' for the event object. Another type of augmentation comes from the action tokens. For example, G2 utilizes ``tell me'' for the ``Query'' action in the intent of ``Query Alarm'', while G5 generates ``check'' for recommendation. 
There are also bad cases like B3 that is generated for ``Query Alarm'' but comes from a similar intent ``Set Alarm''. The other type of bad case, like B6, has syntax errors.
\section{Conclusions}
In this paper, we propose a Pseudo Siamese Network (PSN) to generate labeled data for few-shot intents. PSN consists of two subnetworks (an action network and an object network) with the same structure but different weights. Each sub-network is a transformer-based variational autoencoder. They are trained to learn either the action or the object existing in the intent. It provides an interpretable framework for generating an utterance for a given intent. Experiments on two real-world datasets show that PSN achieves state-of-the-art performance for the generalized few shot intent detection task.

\section*{Acknowledgments}
We thank the reviewers for their valuable comments. This work is supported in part by NSF under grants III-1763325, III-1909323, and SaTC-1930941. 

\newpage
\bibliographystyle{ACM-Reference-Format}
\balance
\bibliography{sigir2021}


\begin{thebibliography}{28}


\ifx \showCODEN    \undefined \def \showCODEN     #1{\unskip}     \fi
\ifx \showDOI      \undefined \def \showDOI       #1{#1}\fi
\ifx \showISBNx    \undefined \def \showISBNx     #1{\unskip}     \fi
\ifx \showISBNxiii \undefined \def \showISBNxiii  #1{\unskip}     \fi
\ifx \showISSN     \undefined \def \showISSN      #1{\unskip}     \fi
\ifx \showLCCN     \undefined \def \showLCCN      #1{\unskip}     \fi
\ifx \shownote     \undefined \def \shownote      #1{#1}          \fi
\ifx \showarticletitle \undefined \def \showarticletitle #1{#1}   \fi
\ifx \showURL      \undefined \def \showURL       {\relax}        \fi
\providecommand\bibfield[2]{#2}
\providecommand\bibinfo[2]{#2}
\providecommand\natexlab[1]{#1}
\providecommand\showeprint[2][]{arXiv:#2}

\bibitem[\protect\citeauthoryear{Ba, Kiros, and Hinton}{Ba
  et~al\mbox{.}}{2016}]%
        {ba2016layer}
\bibfield{author}{\bibinfo{person}{Jimmy~Lei Ba}, \bibinfo{person}{Jamie~Ryan
  Kiros}, {and} \bibinfo{person}{Geoffrey~E Hinton}.}
  \bibinfo{year}{2016}\natexlab{}.
\newblock \showarticletitle{Layer normalization}.
\newblock \bibinfo{journal}{\emph{arXiv preprint arXiv:1607.06450}}
  (\bibinfo{year}{2016}).
\newblock


\bibitem[\protect\citeauthoryear{Bowman, Vilnis, Vinyals, Dai, Jozefowicz, and
  Bengio}{Bowman et~al\mbox{.}}{2015}]%
        {bowman2015generating}
\bibfield{author}{\bibinfo{person}{Samuel~R Bowman}, \bibinfo{person}{Luke
  Vilnis}, \bibinfo{person}{Oriol Vinyals}, \bibinfo{person}{Andrew~M Dai},
  \bibinfo{person}{Rafal Jozefowicz}, {and} \bibinfo{person}{Samy Bengio}.}
  \bibinfo{year}{2015}\natexlab{}.
\newblock \showarticletitle{Generating sentences from a continuous space}.
\newblock \bibinfo{journal}{\emph{arXiv preprint arXiv:1511.06349}}
  (\bibinfo{year}{2015}).
\newblock


\bibitem[\protect\citeauthoryear{Chen, Hakkani-T{\"u}r, T{\"u}r, Gao, and
  Deng}{Chen et~al\mbox{.}}{2016}]%
        {chen2016end}
\bibfield{author}{\bibinfo{person}{Yun-Nung Chen}, \bibinfo{person}{Dilek
  Hakkani-T{\"u}r}, \bibinfo{person}{G{\"o}khan T{\"u}r},
  \bibinfo{person}{Jianfeng Gao}, {and} \bibinfo{person}{Li Deng}.}
  \bibinfo{year}{2016}\natexlab{}.
\newblock \showarticletitle{End-to-End Memory Networks with Knowledge Carryover
  for Multi-Turn Spoken Language Understanding.}. In
  \bibinfo{booktitle}{\emph{INTERSPEECH}}. \bibinfo{pages}{3245--3249}.
\newblock


\bibitem[\protect\citeauthoryear{Coucke, Saade, Ball, Bluche, Caulier, Leroy,
  Doumouro, Gisselbrecht, Caltagirone, Lavril, et~al\mbox{.}}{Coucke
  et~al\mbox{.}}{2018}]%
        {coucke2018snips}
\bibfield{author}{\bibinfo{person}{Alice Coucke}, \bibinfo{person}{Alaa Saade},
  \bibinfo{person}{Adrien Ball}, \bibinfo{person}{Th{\'e}odore Bluche},
  \bibinfo{person}{Alexandre Caulier}, \bibinfo{person}{David Leroy},
  \bibinfo{person}{Cl{\'e}ment Doumouro}, \bibinfo{person}{Thibault
  Gisselbrecht}, \bibinfo{person}{Francesco Caltagirone},
  \bibinfo{person}{Thibaut Lavril}, {et~al\mbox{.}}}
  \bibinfo{year}{2018}\natexlab{}.
\newblock \showarticletitle{Snips voice platform: an embedded spoken language
  understanding system for private-by-design voice interfaces}.
\newblock \bibinfo{journal}{\emph{arXiv preprint arXiv:1805.10190}}
  (\bibinfo{year}{2018}).
\newblock


\bibitem[\protect\citeauthoryear{Devlin, Chang, Lee, and Toutanova}{Devlin
  et~al\mbox{.}}{2019}]%
        {devlin2019bert}
\bibfield{author}{\bibinfo{person}{Jacob Devlin}, \bibinfo{person}{Ming-Wei
  Chang}, \bibinfo{person}{Kenton Lee}, {and} \bibinfo{person}{Kristina
  Toutanova}.} \bibinfo{year}{2019}\natexlab{}.
\newblock \showarticletitle{BERT: Pre-training of Deep Bidirectional
  Transformers for Language Understanding}. In
  \bibinfo{booktitle}{\emph{NAACL}}. \bibinfo{pages}{4171--4186}.
\newblock


\bibitem[\protect\citeauthoryear{Dong, Yang, Wang, Wei, Liu, Wang, Gao, Zhou,
  and Hon}{Dong et~al\mbox{.}}{2019}]%
        {dong2019unified}
\bibfield{author}{\bibinfo{person}{Li Dong}, \bibinfo{person}{Nan Yang},
  \bibinfo{person}{Wenhui Wang}, \bibinfo{person}{Furu Wei},
  \bibinfo{person}{Xiaodong Liu}, \bibinfo{person}{Yu Wang},
  \bibinfo{person}{Jianfeng Gao}, \bibinfo{person}{Ming Zhou}, {and}
  \bibinfo{person}{Hsiao-Wuen Hon}.} \bibinfo{year}{2019}\natexlab{}.
\newblock \showarticletitle{Unified Language Model Pre-training for Natural
  Language Understanding and Generation}.
\newblock \bibinfo{journal}{\emph{arXiv preprint arXiv:1905.03197}}
  (\bibinfo{year}{2019}).
\newblock


\bibitem[\protect\citeauthoryear{Hochreiter and Schmidhuber}{Hochreiter and
  Schmidhuber}{1997}]%
        {hochreiter1997long}
\bibfield{author}{\bibinfo{person}{Sepp Hochreiter} {and}
  \bibinfo{person}{J{\"u}rgen Schmidhuber}.} \bibinfo{year}{1997}\natexlab{}.
\newblock \showarticletitle{Long short-term memory}.
\newblock \bibinfo{journal}{\emph{Neural computation}} \bibinfo{volume}{9},
  \bibinfo{number}{8} (\bibinfo{year}{1997}), \bibinfo{pages}{1735--1780}.
\newblock


\bibitem[\protect\citeauthoryear{Hu, Yang, Liang, Salakhutdinov, and Xing}{Hu
  et~al\mbox{.}}{2017}]%
        {hu2017toward}
\bibfield{author}{\bibinfo{person}{Zhiting Hu}, \bibinfo{person}{Zichao Yang},
  \bibinfo{person}{Xiaodan Liang}, \bibinfo{person}{Ruslan Salakhutdinov},
  {and} \bibinfo{person}{Eric~P Xing}.} \bibinfo{year}{2017}\natexlab{}.
\newblock \showarticletitle{Toward controlled generation of text}. In
  \bibinfo{booktitle}{\emph{Proceedings of the 34th International Conference on
  Machine Learning-Volume 70}}. JMLR. org, \bibinfo{pages}{1587--1596}.
\newblock


\bibitem[\protect\citeauthoryear{Kingma, Mohamed, Jimenez~Rezende, and
  Welling}{Kingma et~al\mbox{.}}{2014}]%
        {NIPS2014_5352}
\bibfield{author}{\bibinfo{person}{Durk~P Kingma}, \bibinfo{person}{Shakir
  Mohamed}, \bibinfo{person}{Danilo Jimenez~Rezende}, {and}
  \bibinfo{person}{Max Welling}.} \bibinfo{year}{2014}\natexlab{}.
\newblock \showarticletitle{Semi-supervised Learning with Deep Generative
  Models}.
\newblock In \bibinfo{booktitle}{\emph{Advances in Neural Information
  Processing Systems 27}}, \bibfield{editor}{\bibinfo{person}{Z.~Ghahramani},
  \bibinfo{person}{M.~Welling}, \bibinfo{person}{C.~Cortes},
  \bibinfo{person}{N.~D. Lawrence}, {and} \bibinfo{person}{K.~Q. Weinberger}}
  (Eds.). \bibinfo{publisher}{Curran Associates, Inc.},
  \bibinfo{pages}{3581--3589}.
\newblock
\urldef\tempurl%
\url{http://papers.nips.cc/paper/5352-semi-supervised-learning-with-deep-generative-models.pdf}
\showURL{%
\tempurl}


\bibitem[\protect\citeauthoryear{Kingma and Welling}{Kingma and
  Welling}{2013}]%
        {kingma2013auto}
\bibfield{author}{\bibinfo{person}{Diederik~P Kingma} {and}
  \bibinfo{person}{Max Welling}.} \bibinfo{year}{2013}\natexlab{}.
\newblock \showarticletitle{Auto-encoding variational bayes}.
\newblock \bibinfo{journal}{\emph{arXiv preprint arXiv:1312.6114}}
  (\bibinfo{year}{2013}).
\newblock


\bibitem[\protect\citeauthoryear{Liu, Fan, Wang, and Yu}{Liu
  et~al\mbox{.}}{2021}]%
        {liu2021augmenting}
\bibfield{author}{\bibinfo{person}{Zhiwei Liu}, \bibinfo{person}{Ziwei Fan},
  \bibinfo{person}{Yu Wang}, {and} \bibinfo{person}{Philip~S. Yu}.}
  \bibinfo{year}{2021}\natexlab{}.
\newblock \showarticletitle{Augmenting Sequential Recommendation with
  Pseudo-PriorItems via Reversely Pre-training Transformer}.
\newblock \bibinfo{journal}{\emph{Proceedings of the 44th international ACM
  SIGIR conference on Research and development in information retrieval}}.
\newblock


\bibitem[\protect\citeauthoryear{Malandrakis, Shen, Goyal, Gao, Sethi, and
  Metallinou}{Malandrakis et~al\mbox{.}}{2019}]%
        {malandrakis2019controlled}
\bibfield{author}{\bibinfo{person}{Nikolaos Malandrakis},
  \bibinfo{person}{Minmin Shen}, \bibinfo{person}{Anuj Goyal},
  \bibinfo{person}{Shuyang Gao}, \bibinfo{person}{Abhishek Sethi}, {and}
  \bibinfo{person}{Angeliki Metallinou}.} \bibinfo{year}{2019}\natexlab{}.
\newblock \showarticletitle{Controlled Text Generation for Data Augmentation in
  Intelligent Artificial Agents}.
\newblock \bibinfo{journal}{\emph{arXiv preprint arXiv:1910.03487}}
  (\bibinfo{year}{2019}).
\newblock


\bibitem[\protect\citeauthoryear{Nguyen, Zhang, Xia, and Philip}{Nguyen
  et~al\mbox{.}}{2020}]%
        {nguyen2020semantic}
\bibfield{author}{\bibinfo{person}{Hoang Nguyen}, \bibinfo{person}{Chenwei
  Zhang}, \bibinfo{person}{Congying Xia}, {and} \bibinfo{person}{S~Yu Philip}.}
  \bibinfo{year}{2020}\natexlab{}.
\newblock \showarticletitle{Semantic Matching and Aggregation Network for
  Few-shot Intent Detection}. In \bibinfo{booktitle}{\emph{Proceedings of the
  2020 Conference on Empirical Methods in Natural Language Processing:
  Findings}}. \bibinfo{pages}{1209--1218}.
\newblock


\bibitem[\protect\citeauthoryear{Snell, Swersky, and Zemel}{Snell
  et~al\mbox{.}}{2017}]%
        {snell2017prototypical}
\bibfield{author}{\bibinfo{person}{Jake Snell}, \bibinfo{person}{Kevin
  Swersky}, {and} \bibinfo{person}{Richard Zemel}.}
  \bibinfo{year}{2017}\natexlab{}.
\newblock \showarticletitle{Prototypical networks for few-shot learning}. In
  \bibinfo{booktitle}{\emph{Advances in Neural Information Processing
  Systems}}. \bibinfo{pages}{4077--4087}.
\newblock


\bibitem[\protect\citeauthoryear{Vaswani, Shazeer, Parmar, Uszkoreit, Jones,
  Gomez, Kaiser, and Polosukhin}{Vaswani et~al\mbox{.}}{2017}]%
        {vaswani2017attention}
\bibfield{author}{\bibinfo{person}{Ashish Vaswani}, \bibinfo{person}{Noam
  Shazeer}, \bibinfo{person}{Niki Parmar}, \bibinfo{person}{Jakob Uszkoreit},
  \bibinfo{person}{Llion Jones}, \bibinfo{person}{Aidan~N Gomez},
  \bibinfo{person}{{\L}ukasz Kaiser}, {and} \bibinfo{person}{Illia
  Polosukhin}.} \bibinfo{year}{2017}\natexlab{}.
\newblock \showarticletitle{Attention is all you need}. In
  \bibinfo{booktitle}{\emph{Advances in neural information processing
  systems}}. \bibinfo{pages}{5998--6008}.
\newblock


\bibitem[\protect\citeauthoryear{Wei and Zou}{Wei and Zou}{2019}]%
        {wei2019eda}
\bibfield{author}{\bibinfo{person}{Jason~W Wei} {and} \bibinfo{person}{Kai
  Zou}.} \bibinfo{year}{2019}\natexlab{}.
\newblock \showarticletitle{Eda: Easy data augmentation techniques for boosting
  performance on text classification tasks}.
\newblock \bibinfo{journal}{\emph{arXiv preprint arXiv:1901.11196}}
  (\bibinfo{year}{2019}).
\newblock


\bibitem[\protect\citeauthoryear{Wu, Schuster, Chen, Le, Norouzi, Macherey,
  Krikun, Cao, Gao, Macherey, et~al\mbox{.}}{Wu et~al\mbox{.}}{2016}]%
        {wu2016google}
\bibfield{author}{\bibinfo{person}{Yonghui Wu}, \bibinfo{person}{Mike
  Schuster}, \bibinfo{person}{Zhifeng Chen}, \bibinfo{person}{Quoc~V Le},
  \bibinfo{person}{Mohammad Norouzi}, \bibinfo{person}{Wolfgang Macherey},
  \bibinfo{person}{Maxim Krikun}, \bibinfo{person}{Yuan Cao},
  \bibinfo{person}{Qin Gao}, \bibinfo{person}{Klaus Macherey}, {et~al\mbox{.}}}
  \bibinfo{year}{2016}\natexlab{}.
\newblock \showarticletitle{Google's neural machine translation system:
  Bridging the gap between human and machine translation}.
\newblock \bibinfo{journal}{\emph{arXiv preprint arXiv:1609.08144}}
  (\bibinfo{year}{2016}).
\newblock


\bibitem[\protect\citeauthoryear{Xia, Xiong, Philip, and Socher}{Xia
  et~al\mbox{.}}{2020a}]%
        {xia2020composed}
\bibfield{author}{\bibinfo{person}{Congying Xia}, \bibinfo{person}{Caiming
  Xiong}, \bibinfo{person}{S~Yu Philip}, {and} \bibinfo{person}{Richard
  Socher}.} \bibinfo{year}{2020}\natexlab{a}.
\newblock \showarticletitle{Composed Variational Natural Language Generation
  for Few-shot Intents}. In \bibinfo{booktitle}{\emph{Proceedings of the 2020
  Conference on Empirical Methods in Natural Language Processing: Findings}}.
  \bibinfo{pages}{3379--3388}.
\newblock


\bibitem[\protect\citeauthoryear{Xia, Yin, Feng, and Yu}{Xia
  et~al\mbox{.}}{2021}]%
        {xia2021incremental}
\bibfield{author}{\bibinfo{person}{Congying Xia}, \bibinfo{person}{Wenpeng
  Yin}, \bibinfo{person}{Yihao Feng}, {and} \bibinfo{person}{Philip Yu}.}
  \bibinfo{year}{2021}\natexlab{}.
\newblock \showarticletitle{Incremental Few-shot Text Classification with
  Multi-round New Classes: Formulation, Dataset and System}.
\newblock \bibinfo{journal}{\emph{arXiv preprint arXiv:2104.11882}}
  (\bibinfo{year}{2021}).
\newblock


\bibitem[\protect\citeauthoryear{Xia, Zhang, Nguyen, Zhang, and Yu}{Xia
  et~al\mbox{.}}{2020b}]%
        {xia2020cg}
\bibfield{author}{\bibinfo{person}{Congying Xia}, \bibinfo{person}{Chenwei
  Zhang}, \bibinfo{person}{Hoang Nguyen}, \bibinfo{person}{Jiawei Zhang}, {and}
  \bibinfo{person}{Philip Yu}.} \bibinfo{year}{2020}\natexlab{b}.
\newblock \showarticletitle{CG-BERT: Conditional Text Generation with BERT for
  Generalized Few-shot Intent Detection}.
\newblock \bibinfo{journal}{\emph{arXiv preprint arXiv:2004.01881}}
  (\bibinfo{year}{2020}).
\newblock


\bibitem[\protect\citeauthoryear{Xia, Zhang, Yan, Chang, and Philip}{Xia
  et~al\mbox{.}}{2018}]%
        {xia2018zero}
\bibfield{author}{\bibinfo{person}{Congying Xia}, \bibinfo{person}{Chenwei
  Zhang}, \bibinfo{person}{Xiaohui Yan}, \bibinfo{person}{Yi Chang}, {and}
  \bibinfo{person}{S~Yu Philip}.} \bibinfo{year}{2018}\natexlab{}.
\newblock \showarticletitle{Zero-shot User Intent Detection via Capsule Neural
  Networks}. In \bibinfo{booktitle}{\emph{Proceedings of the 2018 Conference on
  Empirical Methods in Natural Language Processing}}.
  \bibinfo{pages}{3090--3099}.
\newblock


\bibitem[\protect\citeauthoryear{Xia, Zhang, Zhang, Liang, Peng, and Yu}{Xia
  et~al\mbox{.}}{2020c}]%
        {9319339}
\bibfield{author}{\bibinfo{person}{Congying Xia}, \bibinfo{person}{Chenwei
  Zhang}, \bibinfo{person}{Jiawei Zhang}, \bibinfo{person}{Tingting Liang},
  \bibinfo{person}{Hao Peng}, {and} \bibinfo{person}{Philip~S. Yu}.}
  \bibinfo{year}{2020}\natexlab{c}.
\newblock \showarticletitle{Low-shot Learning in Natural Language Processing}.
  In \bibinfo{booktitle}{\emph{2020 IEEE Second International Conference on
  Cognitive Machine Intelligence (CogMI)}}. \bibinfo{pages}{185--189}.
\newblock
\urldef\tempurl%
\url{https://doi.org/10.1109/CogMI50398.2020.00031}
\showDOI{\tempurl}


\bibitem[\protect\citeauthoryear{Xingkun~Liu and Rieser}{Xingkun~Liu and
  Rieser}{2019}]%
        {XLiu.etal:IWSDS2019}
\bibfield{author}{\bibinfo{person}{Pawel~Swietojanski Xingkun~Liu,
  Arash~Eshghi} {and} \bibinfo{person}{Verena Rieser}.}
  \bibinfo{year}{2019}\natexlab{}.
\newblock \showarticletitle{Benchmarking Natural Language Understanding
  Services for building Conversational Agents}. In
  \bibinfo{booktitle}{\emph{Proceedings of the Tenth International Workshop on
  Spoken Dialogue Systems Technology (IWSDS)}}. \bibinfo{publisher}{Springer},
  \bibinfo{address}{Ortigia, Siracusa (SR), Italy}, \bibinfo{pages}{xxx--xxx}.
\newblock
\urldef\tempurl%
\url{http://www.xx.xx/xx/}
\showURL{%
\tempurl}


\bibitem[\protect\citeauthoryear{Xu, Liu, Shu, and Yu}{Xu
  et~al\mbox{.}}{2019}]%
        {xu2019open}
\bibfield{author}{\bibinfo{person}{Hu Xu}, \bibinfo{person}{Bing Liu},
  \bibinfo{person}{Lei Shu}, {and} \bibinfo{person}{P Yu}.}
  \bibinfo{year}{2019}\natexlab{}.
\newblock \showarticletitle{Open-world Learning and Application to Product
  Classification}. In \bibinfo{booktitle}{\emph{The World Wide Web
  Conference}}. ACM, \bibinfo{pages}{3413--3419}.
\newblock


\bibitem[\protect\citeauthoryear{Xu and Sarikaya}{Xu and Sarikaya}{2013}]%
        {xu2013convolutional}
\bibfield{author}{\bibinfo{person}{Puyang Xu} {and} \bibinfo{person}{Ruhi
  Sarikaya}.} \bibinfo{year}{2013}\natexlab{}.
\newblock \showarticletitle{Convolutional neural network based triangular crf
  for joint intent detection and slot filling}. In
  \bibinfo{booktitle}{\emph{ASRU}}. \bibinfo{pages}{78--83}.
\newblock


\bibitem[\protect\citeauthoryear{Yoo, Shin, and Lee}{Yoo et~al\mbox{.}}{2019}]%
        {yoo2019data}
\bibfield{author}{\bibinfo{person}{Kang~Min Yoo}, \bibinfo{person}{Youhyun
  Shin}, {and} \bibinfo{person}{Sang-goo Lee}.}
  \bibinfo{year}{2019}\natexlab{}.
\newblock \showarticletitle{Data augmentation for spoken language understanding
  via joint variational generation}. In \bibinfo{booktitle}{\emph{Proceedings
  of the AAAI Conference on Artificial Intelligence}},
  Vol.~\bibinfo{volume}{33}. \bibinfo{pages}{7402--7409}.
\newblock


\bibitem[\protect\citeauthoryear{Zhang, Hashimoto, Liu, Wu, Wan, Philip,
  Socher, and Xiong}{Zhang et~al\mbox{.}}{2020}]%
        {zhang2020discriminative}
\bibfield{author}{\bibinfo{person}{Jianguo Zhang}, \bibinfo{person}{Kazuma
  Hashimoto}, \bibinfo{person}{Wenhao Liu}, \bibinfo{person}{Chien-Sheng Wu},
  \bibinfo{person}{Yao Wan}, \bibinfo{person}{S~Yu Philip},
  \bibinfo{person}{Richard Socher}, {and} \bibinfo{person}{Caiming Xiong}.}
  \bibinfo{year}{2020}\natexlab{}.
\newblock \showarticletitle{Discriminative Nearest Neighbor Few-Shot Intent
  Detection by Transferring Natural Language Inference}. In
  \bibinfo{booktitle}{\emph{Proceedings of the 2020 Conference on Empirical
  Methods in Natural Language Processing (EMNLP)}}.
  \bibinfo{pages}{5064--5082}.
\newblock


\bibitem[\protect\citeauthoryear{Zhao, Zhao, and Eskenazi}{Zhao
  et~al\mbox{.}}{2017}]%
        {zhao2017learning}
\bibfield{author}{\bibinfo{person}{Tiancheng Zhao}, \bibinfo{person}{Ran Zhao},
  {and} \bibinfo{person}{Maxine Eskenazi}.} \bibinfo{year}{2017}\natexlab{}.
\newblock \showarticletitle{Learning Discourse-level Diversity for Neural
  Dialog Models using Conditional Variational Autoencoders}. In
  \bibinfo{booktitle}{\emph{Proceedings of the 55th Annual Meeting of the
  Association for Computational Linguistics (Volume 1: Long Papers)}}.
  \bibinfo{pages}{654--664}.
\newblock


\end{thebibliography}



\end{document}